\documentclass{article}

     \usepackage[preprint, nonatbib]{tackling_climate_workshop_style}

\usepackage[utf8]{inputenc} 
\usepackage[T1]{fontenc}    
\usepackage{hyperref}       
\usepackage{url}            
\usepackage{float}
\usepackage{adjustbox}

\usepackage{booktabs}       
\usepackage{amsfonts}       
\usepackage{nicefrac}       
\usepackage{microtype}      
\usepackage{graphicx}
\graphicspath{ {images/} }
\usepackage{setspace}
\usepackage{pgfplots} 
\usepackage{caption} 
\usepackage{booktabs}
\usepackage{biblatex}

\bibliography{bibfile}

\title{\textit{Predicting Tornadoes days ahead with Machine Learning}}

\author{ Davide Alessandro Coccomini \\
        ISTI-CNR \\
        Via G. Moruzzi 1, 56124, Pisa, Italy \\
        \texttt{davidealessandro.coccomini@isti.cnr.it}
        \AND
        Giuliano Zara\\
        Genius Loci\\
        Via Carducci 60, 56017, San Giuliano Terme, Pisa \\
        \texttt{zara@geniusloci.tech}
}

\begin{document}

\maketitle

\begin{abstract}
Developing methods to predict disastrous natural phenomena is more important than ever, and tornadoes are among the most dangerous ones in nature. Due to the unpredictability of the weather, counteracting them is not an easy task and today it is mainly carried out by expert meteorologists, who interpret meteorological models. In this paper we propose a system for the early detection of a tornado, validating its effectiveness in a real-world context and exploiting meteorological data collection systems that are already widespread throughout the world. Our system was able to predict tornadoes with a maximum probability of 84\% up to five days before the event on a novel dataset of more than 5000 tornadic and non-tornadic events. The dataset and the code to reproduce our results are available at: \url{https://tinyurl.com/3brsfwpk}
\end{abstract}

\section{Introduction}

Tornadoes are today one of the most destructive and frightening natural phenomena in the planet. On average, the economic damage caused by tornadoes in the US only is estimated at USD 3.1 billion per year \footnote{https://www.statista.com/statistics/237409/economic-damage-caused-by-tornadoes-in-us/}. Getting worse the situation is climate change. Some studies predict that it may generate conditions for more severe thunderstorms by increasing the opportunities for tornadoes to form \footnote{https://education.nationalgeographic.org/resource/tornadoes-and-climate-change}. Fighting the causes that generate natural disasters with increasing frequency and intensity undoubtedly is the best strategy, but doing that requires international agreements which are often not easy to conclude. Thus, the most realistic and immediate goal at the moment is to predict a tornado sufficiently in advance to be able to save lives and drastically reduce economic damage. The tornadogenesis, however, is a very chaotic process \cite{https://doi.org/10.1002/met.1437} which makes prediction very hard, also because of an incomplete understanding of it \cite{Markowski2009TornadogenesisOC}. Nowadays, their prediction is mainly carried out by specialized meteorologists who observe the weather variations in a given area and try to early catch the tornadogenesis \cite{NSSL_TF}. However, this is a very limited approach, which can't be used on a large scale and that is not always enough accurate. It would be therefore desirable to develop automated systems capable of effectively analysing meteorological data and predicting the formation of a tornado as early as possible. Today such systems can be developed by exploiting modern machine learning (ML) techniques which can learn the weather conditions leading to a tornado from an amount of historical examples of meteorological data. Although ML techniques stand out for their effectiveness in countless sectors, they require large amounts of data which unfortunately are not available for the tornadogenesis. 
In this paper, we therefore present a system for the tornadogenesis early detection based on existing and widespread climate metrics collection systems. This allowed us to validate multiple ML techniques on a large amount and realistic data and to produce a proposal for a more sustainable and economically applicable real-world system.

\section{Related Works}
In the last few years, a lot of effort has been made by researchers to tackle the problem of predicting tornadoes. Ji Qiang \cite{qiang2020proposal} for example, proposed a system based on the use of clusters of balloons able to monitor wind speed and also counteract tornadogenesis by reducing the wind speed of the tornado, blocking it and destroying the convective flow of air. However, such methods require a large number of very specific physical instrumentations, which need to be deployed in the risk areasa and this is very expensive and not always feasible. Further, it would also require periodic maintenance or replacement costs. 

Gradually, more and more methods related to ML have been proposed, mainly using Support Vector Machines (SVMs) as in Adrianto et al. \cite{adrianto2009support} where ML has been exploited for predicting the location and time of tornadoes using a set of 33 storm days. Also Trafalis et al. \cite{trafalis2011linear} proposed a rule-based classifier trained with meteorological data collected from 1562 observations (half of them representing tornadoes).

Available datasets for tornado prediction are generally small and unbalanced. This is a rather common situation when working with tasks trying to identify a rare event \cite{Pang_2021}. Trafalis et al. \cite{trafalis2014machine} proposed the most significant dataset in the literature composed by 10816 observed circulations, with 721 of them representing tornadoes. A circulation is a particular weather condition that can lead to the formation of a tornado. For each observation there are 83 attributes such as temperature and wind speed, which are metrics also taken into account in our research. Although in this dataset the number of tornado circulations is quite small, thus making it difficult to train and validate a suitable model. Recently, Aleskerov et al. \cite{aleskerov2020superposition} have proposed, using the same dataset, the application of superposition principle, stating that for all linear systems the net response caused by two or more stimuli is the sum of the responses that would have been caused by each stimulus individually, to tornado prediction, obtaining good results compared to other more traditional solutions. 

Compared to what has been presented in the literature, we propose a more accurate ML-based system capable of performing tornado prediction earlier and also exploiting data-streams from satellites that are already well-distributed around the planet, thus lowering deployment costs.

\section{Method}
\begin{figure}[H]
    \centering
    \includegraphics[width=1\textwidth]{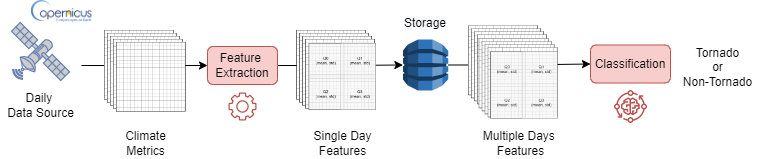}
    \caption{Proposed Tornado Early Detection System overview.}
    \label{figure:system}
\end{figure}

 As shown in Figure \ref{figure:system}, the system exploits meteorological data collected daily from deployed instrumentation which monitors vast geographical areas. 
 In particular, we used as data source the instrumentations of Copernicus\footnote{https://www.copernicus.eu/it}, an initiative of the European Space Agency and the European Commission that makes publicly available the data collected by a series of satellites monitoring the entire surface of the Earth. The data considered by our system are collected over a specific geographical area which is represented by a grid.

\begin{table}[h]
\centering

\begin{tabular}{@{}lrr@{}}
\toprule
\midrule
Name & Unit & Range \\
\midrule
 Temperature          & $K$ &  $[0, +\infty]$ \\
 Wind Components & $m*s^{-1}$ &  $[0, +\infty]$ \\
 Precipitation          &  $m$ & $[0, +\infty]$ \\
 Column Rain Water          & $kg*m^{-2}$ &  $[0, +\infty]$ \\
 Large scale rain rate          &  $kg*m^{-2}*s^{-1}$ &   $[0,
+\infty]$\\
 Cloud Coverage        &  N/A & $[0, 1]$ \\
 
\bottomrule
\bottomrule
\end{tabular}
\caption{Meteorological metrics present in the dataset indicating units of measurement and range.}
\label{tab:metrics}
\end{table}
 
 For each cell of this grid, the meteorological information listed in Table \ref{tab:metrics} are considered. Each grid covers an area of 5 latitude/longitude degrees per side. The considered area is deliberately very large because the process of tornadogenesis can be influenced or triggered by large-scale meteorological phenomena. Further, to predict tornado formation several days in advance is important to monitor areas that are not necessarily close to where the tornadogenesis will occur.  Subsequently, the grid is divided into four quadrants and for each of them, we compute the mean and standard deviation of the metrics values collected for each cell, from now on called features.  Data are daily stored on a database and then the collected meteorological information of several days are used to obtain the outcome from a ML classification model. In order to achieve the earliest possible detection, the system is designed to estimate the tornadogenesis, considering the meteorological metrics in a specific geographical area, up to five days before.

\section{Experiments}
\subsection{Dataset}
In order to compare the performance of various classification techniques and to choose the most suitable one to include in the proposed system, we created a specific dataset. It contains two types of event instances, tornadoes and non-tornadoes. The considered meteorological data source used is the ERA5 dataset \cite{ERA5_DB} representing a worldwide grid with resolution of $0.25\times0.25$ latitude/longitude degrees. 
To label the event instances we used the historical occurrences of tornadoes in the United States collected by the National Centers for Environmental Information \cite{NOAA_DB}. Our dataset consists of 2470 tornadic and 2633 non-tornadic instances and the time period in which these events occur is between 1990 and 2017. Non-tornadic instances are selected by considering areas where tornado events are occurred but in a time period at least 10 days away from any tornadic event in the dataset. For each event, there is weather information for five days. These information are represented as five grids containing all the metrics shown in Table \ref{tab:metrics} and each grid has a size of 19x19 cells. For training the classifiers, the grids values were transformed into features, as explained in Section 2.

\subsection{Results}
In a real-world context, the classifier could be trained on historical tornado data and then have to make predictions on data collected in real time. To simulate this situation, the dataset was divided into training set containing all events collected from 1990 to 2016 and test set composed of 2017 events only. The first one is then composed of 2360 tornadic events and 2633 non-tornadic events while the test set contains 110 tornadic events and 90 non-tornadic events.
Furthermore, in order to investigate the influence of the number of days considered, tests were carried out using as input data to the classifier to make the prediction at day $i$, only the data at day $i-1$, then those at day $i-1$ and $i-2$ and so on until all five daily data available in the dataset were considered. Classifiers, evaluated using two quality metrics commonly used in this field,  Probability Of Detection, $POD = \frac{tp}{(tp+fn)}$ and False Alarm Rate, $FAR =\frac{fp}{(tp+fp)}$ are listed in Table \ref{tab:classifiers}. These metrics are indicative for our system as they measure its ability to generate an alert in the presence of a real tornado and on the other hand not to issue false alarms too frequently.

\begin{table}[]
\resizebox{\textwidth}{!}{
\begin{tabular}{l|cc|cc|ll|ll|ll}
\toprule \midrule
\multicolumn{1}{c|}{Classifier} & \multicolumn{2}{c|}{5 Days} & \multicolumn{2}{c|}{4 Days} & \multicolumn{2}{c|}{3 Days} & \multicolumn{2}{c|}{2 Days} & \multicolumn{2}{c}{1 Day} \\ \midrule
                                 & POD     & FAR         & POD     & FAR         & POD     & FAR     & POD     & FAR     & POD    & FAR    \\ \midrule
Gaussian Classifier                 & 0.79    & 0.10       & 0.83    & 0.05      & 0.80    & 0.06      & 0.75    & 0.07      & 0.61   & 0.07   \\ 
Decision Tree                       & 0.59    & 0.31       & 0.55    & 0.41        & 0.58    & 0.32       & 0.59    & 0.34    & 0.51   & 0.55   \\ 
\textbf{Random Forest}                      & \textbf{0.84}    & \textbf{0.06}       & \textbf{0.84}    & \textbf{0.07}      & \textbf{0.77}    & \textbf{0.09}     & \textbf{0.62}    & \textbf{0.20}     & \textbf{0.76}   & \textbf{0.10}   \\ 
SVM                                & 0.70    & 0.25     & 0.72    & 0.21     & 0.63    & 0.32      & 0.60    & 0.38       & 0.54   & 0.32   \\ 
K-nearest Neighors Classifier     & 0.62    & 0.28       & 0.60    & 0.24       & 0.59    & 0.36       & 0.61    & 0.31      & 0.62   & 0.27   \\ 
AdaBoost Classifier                & 0.64    & 0.29       & 0.68    & 0.24       & 0.63    & 0.36      & 0.61    & 0.25      & 0.62   & 0.25   \\ \bottomrule \bottomrule
\end{tabular}
}
\caption{Comparison of different classifiers on our dataset}
\label{tab:classifiers}
\end{table}

As can be seen in the table, using four- or five-days data almost always leads to a higher probability of tornado prediction. The Random Forest classifier stands out, which obtains a 84\% POD and a 0.06\% FAR, five days before the tornadogenesis.

By comparing our system, based on Random Forest, with other works in the literature, it can be seen that with this approach we are able to predict a tornado with greater probability and anticipation and with fewer false alarms, as can be seen in the table \ref{tab:sota}

\begin{table}[]
\centering
\begin{tabular}{@{}llll@{}}
\toprule
\midrule
Approach & POD & FAR & Advance  \\
\midrule
Logistic regression \cite{aleskerov2016constructing} & 0.7 & 0.25 & 20 minutes \\
Random forest \cite{aleskerov2016constructing} & 0.58 & 0.17 & 20 minutes \\
Superposition \cite{aleskerov2016constructing} & 0.68 & 0.16 & 20 minutes \\
SVM with threshold adj. \cite{trafalis2014machine} & 0.73 & 0.18 & Present \\
Rotation Forest \cite{trafalis2014machine} & 0.76 & 0.08 & Present \\
\textbf{Random Forest (Our)} & \textbf{0.84} & \textbf{0.06} & \textbf{5 days}  \\
\bottomrule
\bottomrule
\end{tabular}
\caption{Comparison of our best model with previous works. The "Advance" column indicates how far in advance tornadogenesis is predicted.}
\label{tab:sota}
\end{table}

\section{Conclusions}
In this research, we presented a novel early-detection tornado system that, according to our experiments, would be able to identify a tornado with a probability of 84\% and up to five days in advance, validated in a real-world context. Our system can be used to monitor weather conditions in at-risk areas in an automated way and warn authorities early so that action can be taken to stem the massive damage that such disasters can cause. Also of central importance is the sustainability and applicability of the proposed system, which makes use of satelittes that are already widely spread throughout the planet and doesn't require any significant additional expenditure to put it into operation.
The contribution made by this research was also to create a class-balanced and large dataset containing tornadoes and other weathers, paving the way for the development of other tornado forecasting approaches. Thanks to the continued enlargement of chosen data sources, the dataset can also be easily expanded in the future.
It is essential for the scientific community to continue research in this area by proposing novel tornadogenesis early-detection systems to obtain increasingly accurate predictions of these phenomena, thus enabling society to be prepared to tackle this serious and pressing problem, which will be increasingly aggravated by climate change, and prevent disasters.

\printbibliography

\end{document}